\documentclass[conference]{IEEEtran}
\usepackage{graphicx} % Required for inserting images

\IEEEoverridecommandlockouts
\usepackage{cite}
\usepackage{amsmath,amssymb,amsfonts}
\usepackage{algorithm} % For the algorithm floating environment
\usepackage{algpseudocode} % For pseudocode typesetting
\usepackage{graphicx}
\usepackage{textcomp}
\usepackage{xcolor}
\usepackage{multirow}
\usepackage{amssymb}
\usepackage{pifont}
\usepackage{subcaption}
\usepackage{xspace}
\usepackage{bbm}
\usepackage{xspace}
\usepackage{tcolorbox}
\usepackage{color}
\usepackage{framed}
\usepackage{changepage}
\definecolor{shadecolor}{rgb}{0.92,0.92,0.92}
\usepackage{enumitem}

\def\BibTeX{{\rm B\kern-.05em{\sc i\kern-.025em b}\kern-.08em
    T\kern-.1667em\lower.7ex\hbox{E}\kern-.125emX}}
\begin{document}

\title{RED: A Systematic Real-Time Scheduling Approach for Robotic Environmental Dynamics
}

\author{Zexin Li$^{1}$
\xspace\xspace\xspace    Tao Ren$^{1}$
\xspace\xspace\xspace    Xiaoxi He$^{2}$
\xspace\xspace\xspace    Cong Liu$^{1}$\\
$^{1}$University of California, Riverside\\
$^{2}$University of Macau\\
{\tt\small \{zli536, tren013, congl\}@ucr.edu, hexiaoxi@um.edu.mo} \\\
}

\maketitle

\def \Approach{RED\xspace }
\def \tool{RED\xspace }
\newcommand{\zx}[1]{\textcolor{blue}{(zx: #1})}

\begin{abstract}

Intelligent robots are designed to effectively navigate dynamic and unpredictable environments laden with moving mechanical elements and objects. Such environment-induced dynamics, including moving obstacles, can readily alter the computational demand (e.g., the creation of new tasks) and the structure of workloads (e.g., precedence constraints among tasks) during runtime, thereby adversely affecting overall system performance. This challenge is amplified when multi-task inference is expected on robots operating under stringent resource and real-time constraints.
To address such a challenge, we introduce \tool, a systematic real-time scheduling approach designed to support multi-task deep neural network workloads in resource-limited robotic systems. It is designed to adaptively manage the Robotic Environmental Dynamics (RED) while adhering to real-time constraints. At the core of \tool lies a deadline-based scheduler that employs an intermediate deadline assignment policy, effectively managing to change workloads and asynchronous inference prompted by complex, unpredictable environments. This scheduling framework also facilitates the flexible deployment of MIMONet (multi-input multi-output neural networks), which are commonly utilized in multi-tasking robotic systems to circumvent memory bottlenecks.
Building on this scheduling framework, \tool recognizes and leverages a unique characteristic of MIMONet: its weight-shared architecture. To further accommodate and exploit this feature, \tool devises a novel and effective workload refinement and reconstruction process. This process ensures the scheduling framework's compatibility with MIMONet and maximizes efficiency.
We have implemented \tool on several widely used embedded and mobile platforms, including the NVIDIA Jetson Nano, TX2, Xavier, and Orin platforms. We evaluated its performance using workloads that span a broad range of settings typical in navigation robots. The experimental results demonstrate that \tool surpasses existing approaches (often by a significant margin) across critical metrics such as throughput, timing correctness, interference robustness, adaptability, and overhead.

\end{abstract}

%-------------------------------------------------------------------------------

\section{Introduction}

For complex cognitive analysis, intelligent robotic applications increasingly rely on co-executing \textit{multiple} deep neural networks (DNNs), often for \textit{correlated} tasks, known as multi-task inference \cite{bib:MobiCom18:Fang, bib:MobiSys17:Mathur, bib:RTSS19:Xiang, kwon2022xrbench}. These DNNs, each trained for a specific inference task, are often run simultaneously on resource-constrained robotic platforms. Examples include personal robots that recognize sounds and places \cite{bib:MobiSys20:Lee}, autonomous vehicles that see the world from the front, side, and rear views \cite{bib:RTSS19:Xiang}, and home hubs that recognize emotions through both speech and facial expression \cite{bib:ICMR16:Zhang, bib:TMC21:He}.

One imminent challenge for performing multi-task inference in robotic systems is their strict resource and real-time constraints~\cite{xu2022dpmpc,zhang2022learning,botros2022fully,maccio2022mixed,lee2022towards,d3,gog2021pylot}. 
As modern deep learning algorithms are notoriously memory- and computation-hungry, it is challenging to deploy multiple DNNs on resource-constrained hardware platforms while satisfying strict real-time constraints.
Weight sharing across deep neural networks is a class of newly developed methods for increasing drastically the efficiency of multi-task inference \cite{bib:IJCAI18:Chou, bib:NIPS18:He, bib:TMC21:He, bib:MobiSys20:Lee, bib:CVPRW19:Wan, bib:APSIPA20:Wu}. Memory usage can be reduced effectively as parameters are shared across DNNs, which is only possible when these DNNs are intended for correlated inference tasks. However, task correlation is commonplace as many DNNs in multi-task inference use the same input to infer distinct labels (single-input, multi-output) (SIMONet), merge complementary inputs to output a single label jointly (multi-input, single-output) (MISONet), or even a combination of both (multi-input, multi-output) (MIMONet).
% For example, a recent work~\cite{li2023mimonet} adopts MIMONet for a robotic application that accepts both visual and auditory inputs to conduct emotion classification and gender identification.

Intelligent robotic applications often face another challenge: \textit{environmental dynamics}.
Intelligent robots expect to effectively negotiate dynamic, unpredictable environments crowded with moving mechanical elements and objects. Environment-induced dynamics such as moving obstacles could easily transform the computing demand (e.g., new tasks) and structure of workloads (e.g., precedence constraints among tasks) during runtime,  negatively impacting overall system performance. Existing solutions rely on runtime scheduling to cope with the changing environment and runtime variations~\cite{d3,gog2021pylot}. However, as illustrated in Sec. \ref{sec:motivation}, the problem herein is rather challenging due to (\textit{i}) dynamically changing workload (Sec.~\ref{sec:case_study_1}), (\textit{ii}) MIMONet architecture (Sec.~\ref{sec:mimonet_case}), (\textit{iii}) asynchronous inference (Sec.~\ref{sec:case_study_2}). 

To this end, we propose \tool, a systematic real-time scheduling approach designed to enable multi-task DNN workloads in resource-constrained robotic systems that can adaptively handle environmental dynamics and satisfy real-time constraints (i.e., end-to-end deadline of an operation). The center of \tool is a deadline-based scheduler leveraging an intermediate deadline assignment policy, which could efficiently handle changing workloads and asynchronous inference challenges. This scheduling framework also flexibly supports MIMONet architecture (multi-input multi-output neural nets)~\cite{li2023mimonet} for memory-efficient deployment. 
 
Leveraging the MIMONet-driven scheduling framework mentioned above appears to be a feasible and practical solution. However, a critical observation we had is that such a general scheduling framework does not consider the unique properties of MIMONet and may yield suboptimal performance. Specifically, MIMONet features a weight-shared architecture which can be exploited to reduce computation overhead further but may introduce new runtime optimization challenges.

To consider and exploit this feature unique to MIMONet, \tool develops a novel workload refinement and reconstruction process to make the scheduling framework compatible with MIMONet and fully explore its efficiency. An on-demand synchronization is integrated with this scheduling framework aiming to improve the overall system performance by reducing synchronization overhead.
We implemented \tool on four embedded platforms pervasively used in robotic applications, including NVIDIA Jetson Nano, TX2, AGX Xavier, and AGX Orin, which exhibit different computing capabilities and architectural characteristics. We evaluate \tool under a wide range of workloads and settings that navigation robots may experience.

Our main contributions and results are as follows. 1) To the best of our knowledge, we are the first to enable the effective execution of weight-shared DNNs with deadline-driven schedulers. 2) We designed \tool, a unified system that enables efficient multi-task inference on resource-constrained robotic platforms, dealing with real-time constraints in dynamic environments. This holds the potential for more advanced and efficient robotic applications in the future. 3) Our experimental results demonstrate: 

\begin{itemize}[leftmargin=10px]
    \item \textbf{On-device effectiveness}: The \tool demonstrates significantly lower deadline miss rates by 40.5\%, providing an average of 24.7\% reduction in response time and making a 34.8\% improvement in quality of experience as compared to the EDF baseline. (Refer to Sec.~\ref{sec:overall_effectiveness})
    \item \textbf{Practical usability:} \tool can be seamlessly integrated with widely-used robotic middleware ROS 2 and existing systems like NVIDIA IoT AI~\cite{nvidiaiot}, requiring no extensive modifications. Compared to the EDF scheduler built upon ROS2, \tool provides 32.7\% less response time, and lower deadline miss rates by 67.3\%. (Refer to Sec.~\ref{sec:ros2})
    \item \textbf{Robust adaptability}: \tool is capable of automatic adaptation to varying end-to-end deadline settings and diverse quality of experience requirements within acceptable ranges. (Refer to Sec.~\ref{sec:adaptability})
    \item \textbf{Low overhead}: \tool's efficient design and implementation incurs low offline profiling overhead and runtime overhead. (Refer to Sec.~\ref{sec:overhead})
\end{itemize}

\section{Background and Motivational Case Study}
\label{sec:motivation}

\begin{figure}[!t]
    \centering
    \includegraphics[width=0.45\textwidth]{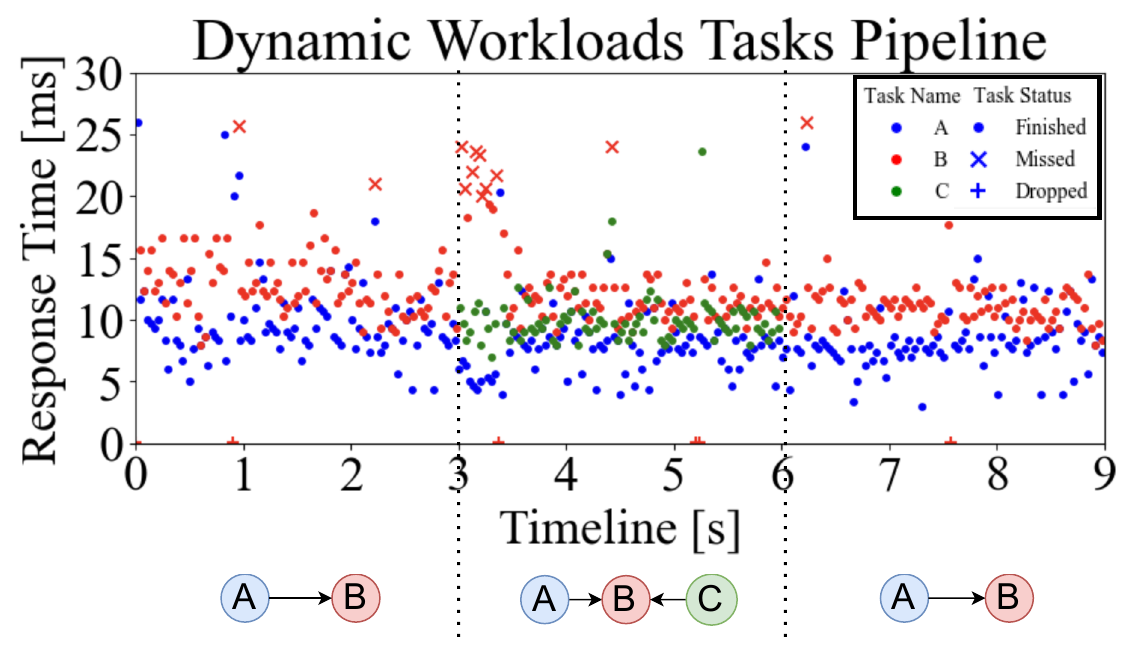}
    \caption{Robotic environmental dynamics induces changes of workloads structures. When environmental dynamics increase, the rate of missing deadlines, tasks dropped rate, and observed execution time of specific tasks dramatically increases, and vice versa.}
    \label{fig:casestudy_1}
\end{figure}

% # -:            parameter   FLOPs
% # encoder:      67,649,097  8.186
% # yolov5:       78,271,872  3.926
% # unet:         7,766,051   2.656
% # enet:         3,529,989   3.665
% # autopilot:    2,959,419   0.143

% \begin{table*}[!htbp]
%   \centering
%   \caption{This table presents an analysis of deep learning-based models utilized in robotic task navigation, along with their corresponding architecture sources. It indicates the specific dataset employed for each model and provides a comparative assessment of memory and GPU memory consumption during the inference stage. }
%   \label{tab:models}
%   \resizebox{0.8\textwidth}{!}{
%     \begin{tabular}{c|c|c|c|c|c|c}
%     \hline 
%     \textbf{Task} & \textbf{Model} & \textbf{Dataset} & \textbf{FLOPs} & \textbf{Params} & \textbf{Mem$_{GPU}$} & \textbf{Mem$_{Process}$} \\
%     \hline
%     Lane Detection & ENet~\cite{paszke2016enet} & KITTI~\cite{geiger2012we} & 11.85B & 71.18M & 562.00MB & 2669.37MB \\
%     Segmentation & UNet~\cite{ronneberger2015u}  & KITTI~\cite{geiger2012we}  & 10.84B & 75.42M & 660.00MB & 2768.07MB \\
%     Cruise Control & AutoPilot~\cite{bojarski2016end}  & Dave~\cite{pan2017virtual}  & 8.33B  & 70.61M & 546.00MB & 2673.46MB \\
%     Object Detection & YOLO~\cite{redmon2016you} & KITTI~\cite{geiger2012we}  & 12.11B & 145.92M &  982.00MB & 2677.94MB \\
%     \hline
%     Multi-tasks & MIMONet & KITTI~\cite{geiger2012we} & 18.58B & 160.18M & 1132.00MB & 2992.12MB \\
%     \hline
%     \end{tabular}
% }
%   \label{tab:succint}
% \end{table*}

\begin{table*}[!htbp]
  \centering
  \caption{This table shows deep learning-based models for navigating robot tasks. The Model column lists the specific models and the sources that describe the model architecture. The dataset column indicates the dataset to be used for each model. For some models, we reduced the resolution of the original model to fit the context of the embedded scenario. We choose YOLOv5n as the minimal version for the object detection task and scale the channel number to 1/4 official UNet.}
  \label{tab:models}
  \resizebox{0.8\textwidth}{!}{
    \begin{tabular}{c|c|c|c|c|c|c}
    % \toprule
    \hline 
    \textbf{Task} & \textbf{Model} & \textbf{Dataset} & \textbf{FLOPs} & \textbf{Params} & \textbf{Mem$_{arm}$} & \textbf{Mem$_{x86}$} \\
    % \midrule
    \hline
    Lane Detection & ENet~\cite{paszke2016enet} & KITTI~\cite{geiger2012we} & 4.111G & 3.666M & 3329MB & 3796MB / 2073MB \\
    Segmentation & UNet~\cite{ronneberger2015u}  & KITTI~\cite{geiger2012we}  & 2.655G & 1.943M & 3359MB & 3752MB / 1261MB \\
    Cruise Control & AutoPilot~\cite{bojarski2016end}  & Dave~\cite{pan2017virtual}  & 0.465G  & 2.489M & 3680MB & 3864MB / 1139MB\\
    Object Detection & YOLO~\cite{redmon2016you} & KITTI~\cite{geiger2012we}  & 0.283G & 5.286M & 3593MB & 3754MB / 1159MB\\
    \hline
    Multi-tasks & MIMONet~\cite{li2023mimonet} & KITTI~\cite{geiger2012we} & 6.704G & 10.465M & 3718MB & 3739MB / 1611MB \\
    \hline
    \end{tabular}%
}
  \label{tab:succint}%
  % \vspace{-4mm}
\end{table*}%

\begin{figure*}[!htbp]
\centering
\includegraphics[width=1.0\textwidth]{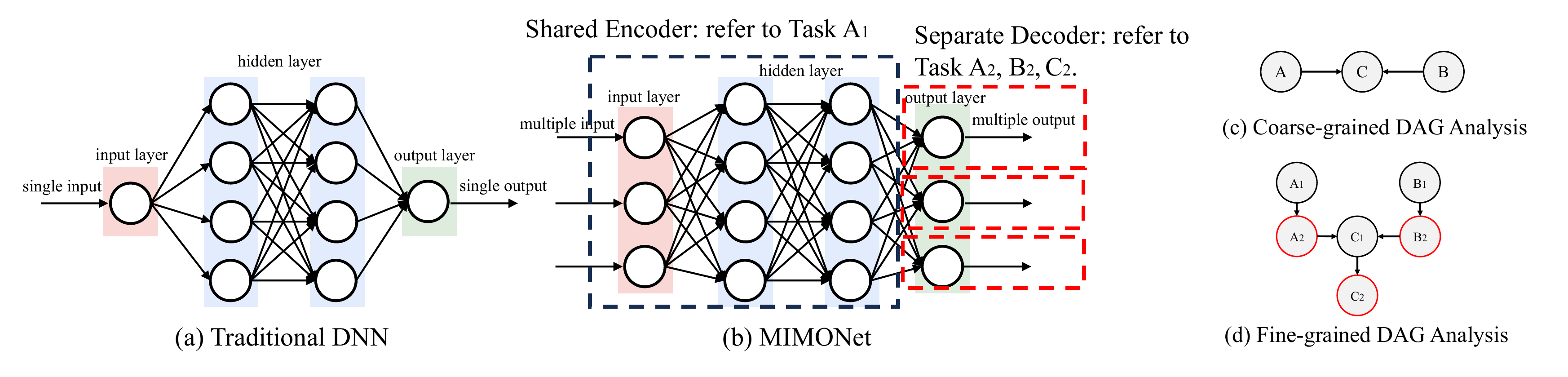}
\caption{The architecture of MIMONet compared to traditional DNN. The structural characteristics of MIMONet (partitionable components) naturally facilitate finer-grained DAG partitioning. Shared encoder inference in (b) marked with a black dashed box refers to a black node in (d), and each separate decoder in (b) marked with a red dashed box refers to a red node in (d).}
\vspace{-5mm}
\label{fig:mimonet_arch}
\end{figure*}

In this section, we present several motivating case studies to examine and comprehend the unique challenges associated with operating multi-input multi-output deep neural network (MIMONet) driven robotic systems. 

\subsection{Challenges due to Dynamically Changing Workloads}
\label{sec:case_study_1}

Robotic systems often function in intricate environments that demand the ability to adapt to dynamically evolving and unpredictable situations. A case in point involves unexpected obstacles, requiring real-time workload adjustments to maintain timing correctness.

A test scenario was established with a refitted TurtleBot3~\cite{bib:turtlebot3} powered by NVIDIA Jetson Nano~\cite{bib:nano}, navigating from point A to point B and back, encountering random obstacles en route. The navigation process, composed of three tasks from the model table, evolved in three phases. Initially, the navigation comprised two dependent tasks: cruising control and segmentation. Upon detecting obstacles, an object detection task was added, causing a substantial increase in deadline misses. When the obstacles cleared, the structure reverted to the initial phase, and the rate of deadline misses declined.

The pivotal challenge, thereby, lies in the dynamic environments where robotic systems operate, inducing changing workload structures. These shifts necessitate real-time adaptive decisions to maintain the end-to-end deadline. The case study data is illustrated in Figure~\ref{fig:casestudy_1} and Table~\ref{tab:models}.

As demonstrated in Figure \ref{fig:casestudy_1}, the navigation process, encapsulating three key tasks from Table~\ref{tab:models}, unfolds in three distinct phases. The initial phase naturally includes two interdependent tasks: cruising control followed by segmentation. Upon encountering an obstacle at a specific juncture (t=3), the object detection task is necessitated, and subsequently incorporated into the workload Directed Acyclic Graph (DAG). Once the environment clears at t=6, the DAG structure reverts to its original configuration.
A, B, and C represent cruising control, segmentation, and object detection. The deadline for A, B, and C is set to 50ms, 30ms, and 50ms, based on their execution times as well as a pre-defined end-to-end deadline of 130ms.

Figure \ref{fig:casestudy_1} reveals a significant escalation in deadline misses during the second phase, coinciding with the unexpected integration of the object detection task into the workload DAG. Specifically, the miss and drop rate for task B surges from 3.3\% to 30.0\%, resulting in an average deadline missing rate of 13.0\% within the interval (3,6). Despite this, the system's GPU remains underutilized during this period. Of note, if task B (the sink node in the DAG) fails to meet its deadline, it leads to a violation of the operation's end-to-end deadline. The removal of task C at t=6 sees a drop in the rate of missed and dropped deadlines for task B to 2.2\%.

\noindent \textbf{Challenge 1: }
Robotic systems frequently operate in dynamic environments that induce fluctuating workload structures. Consequently, robotic system software is required to make real-time adaptive decisions in response to these changing workloads to ensure the end-to-end deadline is met.

\subsection{Challenges due to the MIMONet Architecture }

\label{sec:mimonet_case}

In response to the dynamic challenges posed by robotic systems, advanced computational techniques have become crucial. One area of considerable progress is the evolution of deep neural networks (DNNs). The past decade has witnessed a dramatic expansion of computational capabilities, contributing significantly to the success of DNNs. The field of self-supervised learning-based Large Language Models (LLMs) has notably improved algorithm performance, primarily attributed to their remarkable generalizability, facilitating smooth application across various tasks. Numerous applications founded on these models, including CLIP~\cite{radford2021learning}, DINOv2~\cite{oquab2023dinov2}, GPT-4~\cite{openai2023gpt4}, and Bard~\cite{thoppilan2022lamda}, have attracted significant attention. These applications leverage the core technique of MIMONet (Multiple-Input-Multiple-Output Deep Neural Networks), an advanced DNN architecture enabling simultaneous processing of multiple inputs and outputs for efficient and versatile model training (Figure~\ref{fig:mimonet_arch}). The superior performance of MIMONet is rooted in its ability to decipher complex patterns and dependencies within input data, thereby fostering improved representations and adaptability across a wide spectrum of tasks.

Building on the triumphs of DNNs, the MIMONet architecture has showcased its potential in diverse applications, such as intelligent robots operating within resource-limited embedded systems. Despite variations in implementation, MIMONet architectures manifest common workload characteristics, presenting considerable challenges when deployed in such environments. These characteristics include:

\begin{figure*}[!hbtp]
\centering
\includegraphics[width=\textwidth]{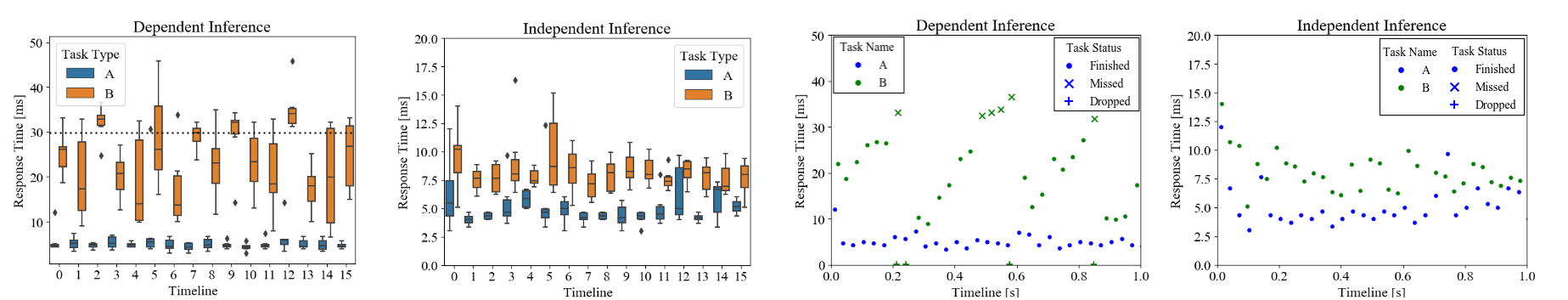}
\caption{Data dependency leads to periodic system's real-time performance degradation in asynchronous inference scenarios. The first two figures present a statistical analysis of end-to-end response times under the two scenarios. The last two figures summarize the task instances that meet, miss, or are dropped by the system.}
\label{fig:casestudy_2}
\vspace{-5mm}
\end{figure*}

\begin{itemize}[leftmargin=10px]
\item \textbf{Memory-Efficient Architecture:} MIMONet is specifically designed to optimize memory usage, making it highly suitable for environments with resource constraints. Even in its simplest form (as depicted in Figure \ref{fig:mimonet_arch}), MIMONet requires only a modest number of additional parameters to facilitate multi-input multi-output support, as compared to traditional DNNs. Table~\ref{tab:models} highlights a selection of critical tasks along with their respective standard models and datasets, underpinning the DNN-driven navigation robots. It is noteworthy that employing MIMONet, as opposed to multiple distinct DNNs, can substantially reduce memory consumption on both ARM-based and x86-based platforms. This observation underscores the memory-efficient design of MIMONet, making it inherently suitable for memory-constrained embedded systems.
\item \textbf{Partitionable Components:} As depicted in Figure~\ref{fig:mimonet_arch}, unlike a traditional DNN (Figure~\ref{fig:mimonet_arch}(a)) that accepts a single input and generates a single output, the network architecture of MIMONet (Figure~\ref{fig:mimonet_arch}(b)) is designed to accept multiple inputs and generate multiple outputs within a single DNN simultaneously. This architectural design boasts a naturally partitionable structure, allowing the inference process to be decomposed into dependent smaller, manageable subtasks that can be scheduled. This feature introduces increased flexibility and adaptability in complex environments, as the subtasks can be dynamically allocated and reconfigured in response to changing environments. 
Figure~\ref{fig:mimonet_arch} further presents a simple showcase to illustrate the potential for MIMONet-enabled optimization based on its structural characteristics. 
This naturally invites finer-grained DAG partitioning while introducing additional optimization challenges, compared to simply MIMONet-agnostic DAG analysis (see Figure~\ref{fig:mimonet_arch}(c)). Specifically, the inference subtask in the black dashed box part, as in Figure~\ref{fig:mimonet_arch}(b), corresponds to the black node of the DAG in Figure~\ref{fig:mimonet_arch}(d) ($A_1, B_1, C_1$), while the inference in the red dashed box corresponds to the red node of the DAG ($A_2, B_2, C_2$). 
\end{itemize}

\noindent \textbf{Challenge 2:} The MIMONet architecture, with its memory-efficient design and partitionable structure, provides a robust solution for intelligent embedded systems. However, it simultaneously presents new runtime optimization challenges that stem from the need to dynamically manage and schedule multiple subtasks in real time in response to changing environmental conditions.

\subsection{Challenges due to Asynchronous Inference}
\label{sec:case_study_2}

In the context of deep learning, asynchronous inference refers to the execution of DNN model inferences without a fixed, synchronized order or timing. Instead of processing data in a strict sequence or at consistent intervals, asynchronous inference allows tasks to be processed based on the availability of resources, task dependencies, or other dynamic factors. This flexibility provides advantages in terms of resource utilization and responsiveness, but it also introduces complexities with respect to task coordination, timing, and system stability.

Specifically, we conduct a case study to attribute complexities in DNN-based robotic systems by introducing asynchronous inference through a two-task asynchronous inference scenario, as shown in Figure~\ref{fig:casestudy_2}.

In this scenario, a TurtleBot3 navigates an obstacle-ridden, hazardous environment, guided by an exploration mission driven by MIMONet systems. The scenario necessitates the constant operation of the object detection module and includes two tasks: Task A, overseeing cruising control, and Task B, managing object detection. Task A operates at a frequency of 30Hz (1/30s) while Task B runs at 33Hz (1/33s), showcasing a subtle variance in frequency data streams. Both tasks have a deadline of 30ms.

This case study explores the implications of asynchronous inference under two conditions: when Tasks A and B are interdependent and when they are not. As Fig.~\ref{fig:casestudy_2} illustrates, in the absence of interdependency, both tasks demonstrate a uniformly distributed performance pattern. However, the introduction of task interdependency results in a significant increase in deadline misses, highlighting the challenges posed by asynchronous inference.

\noindent \textbf{Challenge 3:} The operational framework of a Directed Acyclic Graph (DAG) can spawn asynchronous inference due to task dependencies, potentially causing cascading delays or system failures. This complexity is further intensified when dealing with MIMONet components, which naturally exhibit task interdependence post-partitioning, thereby exacerbating the challenges associated with asynchronous inference. Therefore, it is imperative for robotic system software to incorporate this asynchronicity into runtime resource management decisions.

\section{System Design}
\label{sec:design}

\subsection{Overview}
\label{sec:overview}

\begin{figure*}[!t]
    \centering
    \includegraphics[width=0.8\textwidth] {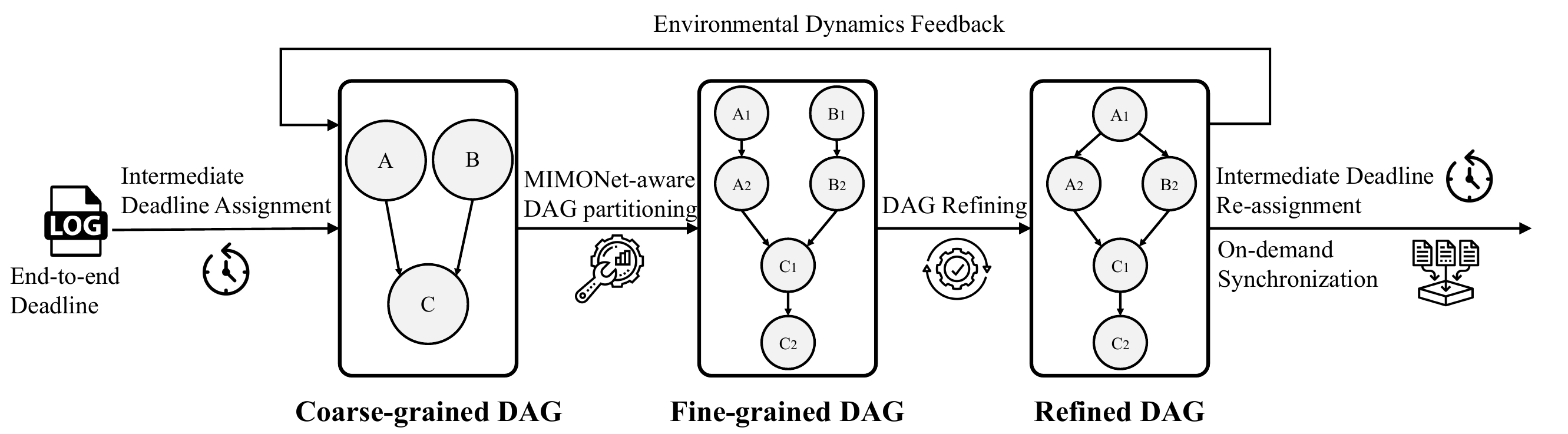}
    \caption{Overview of \tool.}
    \label{fig:overview}
    \vspace{-5mm}
\end{figure*} 

In response to the intricate challenges of ensuring timing correctness for operating robots in complex dynamic environments, as discussed in Sec.~\ref{sec:motivation}, we introduce \tool{}, a comprehensive MIMONet-driven DAG scheduling framework. 
MIMONet inference highly relies on GPU resources. Since most embedded platforms~\cite{bib:agx,bib:nano,bib:tx2,bib:orin} equipped by robots containing one GPU, this study targets single-GPU autonomous embedded platforms, e.g., NVIDIA Jetson Nano~\cite{bib:nano}, TX2~\cite{bib:tx2}, Xavier~\cite{bib:agx}, Orin~\cite{bib:orin}, etc. Note that our proposed method can be further extended to multi-GPU settings.
As illustrated in Figure~\ref{fig:overview}, \tool{} consists of three primary components:
(\textit{i}) A simple yet highly effective intermediate deadline-based scheduler that efficiently adapts to dynamically changing workloads (Challenge 1),
(\textit{ii}) A MIMONet-driven DAG refinement and re-assignment policy that resolves issues arising from the MIMONet structure (Challenge 2), and
(\textit{iii}) An on-demand synchronization mechanism that significantly enhances overall performance under asynchronous inference system scenarios (Challenge 3).
Each component functions in conjunction to address unique challenges and promote optimal system operation, with detailed discussions following below:

\begin{itemize}[leftmargin=10px]
    \item \textbf{Intermediate Deadline Assignment Policy:} The heart of our framework is a robust policy that assigns intermediate deadlines to tasks. This policy is designed to be simple yet highly effective in dealing with the dynamically changing workloads prevalent in complex robotics environments. It ensures efficient scheduling and timely completion of tasks, thereby maintaining the operational correctness of the robots.

    \item  \textbf{MIMONet-Driven DAG Refinement and Deadline Re-Assignment:} To tackle the issues stemming from the MIMONet structure (Challenge 2), our framework incorporates a process of DAG refinement and deadline re-assignment driven by the MIMONet. This component refines the task nodes, adjusting the granularity of tasks and their interconnections according to the weight-shared architecture of MIMONet, which in turn enhances system efficiency against dynamic environmental changes.

    \item  \textbf{On-demand Synchronization:} The last key component of our framework is an on-demand synchronization system aimed at significantly improving the overall performance via largely reducing synchronization overhead. By allowing on-demand synchronization, this system ensures that tasks are executed in harmony with the changing dynamics of the environment, thereby minimizing latency and enhancing the timeliness and effectiveness of robotic operations.
\end{itemize}

Through the seamless integration of the three core components—intermediate deadline assignment policy, MIMONet-driven DAG refinement, and on-demand synchronization system—\tool{} emerges as a high-performance solution to maintain timing correctness for robotic operations in complex environments. As depicted in Figure~\ref{fig:overview}, this framework adroitly handles dynamically changing workloads, resolves challenges intrinsic to the MIMONet structure, and significantly improves overall performance by minimizing synchronization overhead. The holistic design of \tool{}, therefore, offers an efficient system for managing the demands of real-time operations in changing environments, ensuring optimal performance and effective robotic operations

\begin{figure*}[!t]
    \centering
    \includegraphics[width=1.0\textwidth]{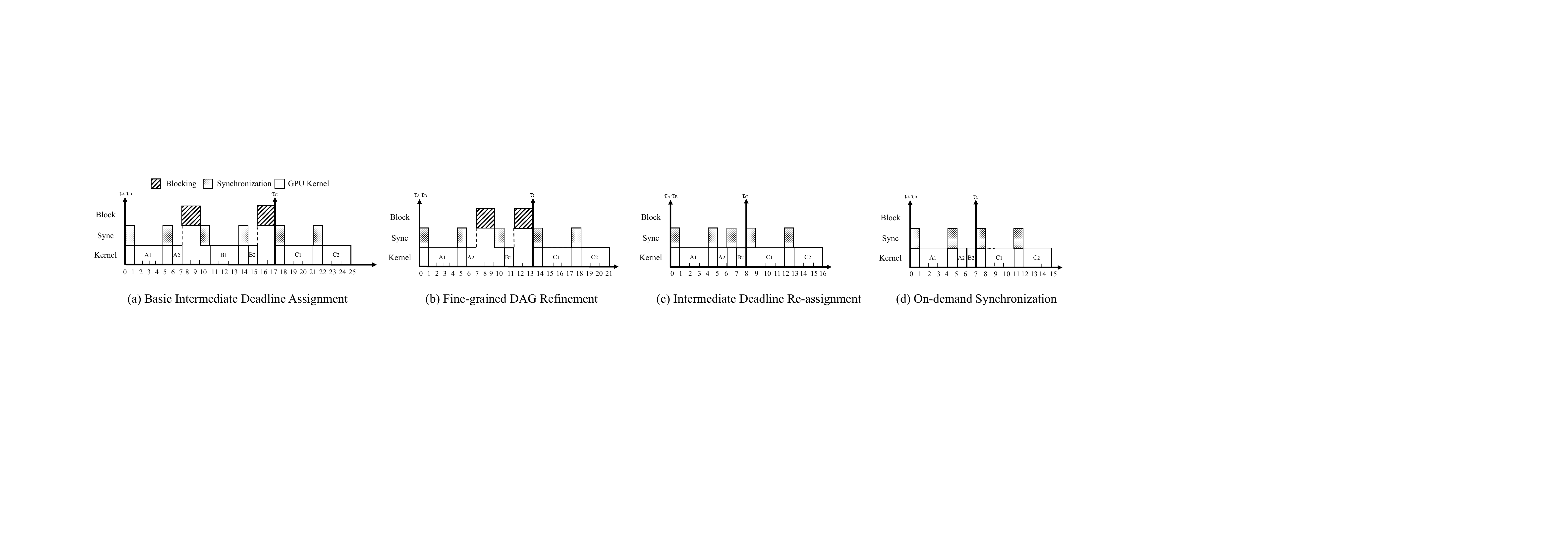}
    \caption{An illustrative example to apply each component of our design to a MIMONet-driven DAG scheduling. \textbf{(a)} Basic intermediate deadline assignment enables incorporation with real-time scheduler and further optimization. \textbf{(b)} Fine-grained DAG refinement reduces computation costs by merging some tasks. \textbf{(c)} Intermediate deadline reassignment reduces blocking time at runtime. \textbf{(d)} On-demand synchronization reduces unnecessary synchronization overhead.}
    \label{fig:finer_grained_dag}
    % \vspace{-2mm}
\end{figure*}

\subsection{Intermediate Deadline Assignment}

Achieving the end-to-end deadline of a robotic operation, represented by a Directed Acyclic Graph (DAG) task, necessitates an efficient deadline assignment strategy. To this end, we propose an intuitive intermediate deadline assignment approach that assigns an intermediate deadline to each node in the DAG task. Essentially, by ensuring each node meets its intermediate deadline, we guarantee the completion of the overall operation within the end-to-end deadline.

%\Cong{This is the first time you introduce these math notations, thus better to integrate the MIMONet info into this describtion so that readers can easily understand the correlation between a MIMONet and its corresponding DAG.}

% \Cong{Add sth to map an MIMONet to a corresponding DAG. Why/how a DAG and its nodes represent an MIMOnet and its components. The system is given a Mimonet workload, not a DAG.}
With the above understanding of MIMONet (as in~\ref{sec:mimonet_case}), we can explore how it interacts with Directed Acyclic Graphs (DAGs) and apply this to a practical example. Consider a DAG $\tau_i$ with a relative end-to-end deadline $D$, comprising $k$ nodes ${ \tau_i^1, \dots, \tau_i^k }$. We assign an intermediate deadline $D_i^k$ to each node. Each node $\tau_i^k$ has a height $H(\tau_i^k)$, representing the length of the longest path from any source node to $\tau_i^k$. By allocating an intermediate deadline $D_i^{H_j}$ to all nodes of height $H_j$ and ensuring $\sum{D_i^{H_j}=D_i}$, the end-to-end-deadline is met if each node in the DAG completes by its assigned intermediate deadline. The partitionable nature of MIMONet's design allows for such tasks to be carried out effectively and efficiently while meeting these intermediate deadlines.

% Consider a DAG $\tau_i$ with an end-to-end deadline $D$, comprising $k$ nodes ${ \tau_i^1, \dots, \tau_i^k }$. We assign an intermediate deadline $D_i^k$ to each node. Each node $\tau_i^k$ has a height $H(\tau_i^k)$, representing the length of the longest path from any source node to $\tau_i^k$. By allocating an intermediate deadline $D_i^{H_j}$ to all nodes of height $H_j$ and ensuring $\sum{D_i^{H_j}=D_i}$, the end-to-end-deadline is met if each node in the DAG completes by its assigned intermediate deadline.

The challenge lies in identifying the optimal intermediate deadline assignment strategy. While equal or proportional assignment to each node's execution time are both intuitive strategies, \tool adopts the latter. Our choice is backed by both empirical evidence from our evaluations and the logical inference that a proportional assignment strategy increases the likelihood of nodes meeting their assigned deadlines. Under an equal assignment strategy, nodes with longer execution times could more easily breach their assigned deadlines.

For instance, consider a coarse-grained DAG shown in Figure~\ref{fig:overview} with an end-to-end deadline of 120ms, where nodes A, B, and C have execution costs of 20ms, 20ms, and 40ms, respectively. Given the dependencies among these nodes, we would assign intermediate deadlines to A, B, and C of 40ms, 40ms, and 80ms, respectively. Thus, if all nodes complete within their intermediate deadlines, the end-to-end deadline is met. Note that a node in the DAG commences execution immediately once all its predecessors have completed. Similarly, in the partitioned DAG in Figure~\ref{fig:finer_grained_dag}\textbf{(a)}, the proportional intermediate deadline can also be assigned directly and work in suboptimal.

While our adoption of proportional intermediate deadline assignment may seem straightforward, it proves to be highly effective and lays the groundwork for the subsequent components of our framework, namely the MIMONet-driven DAG scheduling and on-demand synchronization.

\subsection{MIMONet-driven DAG Scheduling: Refinement and Re-Assignment}

The scheduling of DAG tasks within \tool involves two primary components: fine-grained DAG refinement and MIMONet-driven intermediate deadline reassignment.

\subsubsection{Fine-grained DAG Refinement}
\label{sec:finer_grained_dag_reconstruction}

Weight-shared architectures, such as MIMONet employed in \tool, enable improved system efficiency by sharing parameters across multiple simultaneous tasks as illustrated in Sec.~\ref{sec:mimonet_case}. This structure, comprising a shared encoder followed by several separate decoders, maintains inference accuracy while achieving a smaller overall model size~\cite{caruana1997multitask,maddu2019fisheyemultinet}.

However, this parameter-sharing approach presents unique scheduling challenges. Tasks executed simultaneously can share the intermediate output from the shared encoder, reducing computation costs, as illustrated in Figure~\ref{fig:finer_grained_dag}\textbf{(b)}. Conversely, tasks that start asynchronously cannot share these results, increasing overall computation. It is optimal to delay early tasks slightly to enforce synchronous execution, thereby saving computation and reducing latency. %\zx{Existing ?real time? schedulers, which are MIMONet structure agnostic, may not fully utilize the benefits of MIMONet. }

To address such a challenge, we propose a MIMONet-driven DAG structure refinement. This approach involves performing a finer-grained DAG analysis, leveraging the shared encoder characteristics, thereby significantly reducing computational redundancy and costs. We split each inference sub-task in MIMONet into two dependent parts (encoder inference/decoder inference) and perform semantic-preserving reconstruction of finer-grained DAGs under certain constraints, thus reducing computation cost and taking advantage of parallelism. Algorithm~\ref{alg:algorithm} details this refining process.

Let \( G = (V, E) \) be a topologically sorted directed graph where \( V \) is the set of nodes and \( E \) is the set of edges. Define the indegree of a node \( v \) as:
\[ \text{indegree}(v) = |\{ u \in V : (u, v) \in E \}| \].
The set \( S \) is formed by taking all nodes in \( V \) with indegree($0$) (Line 7). Following that, a dynamic merge operation is conducted (Line 8). The objective of this merge operation is to maximize the efficiency of task execution in a MIMONet architecture, through leveraging the shared encoder characteristics to minimize computational redundancy and costs. Note that only sub-tasks exhibiting release time differences within $\gamma$ will be merged. 

\begin{algorithm}[t!]
\caption{MIMONet DAG structure refining}
\label{alg:algorithm}
\begin{algorithmic}[1]
\State \textbf{Input:}
Task Dependency Graph $G$ 
\State \textbf{Output:}
Fine-grained Task Dependency Graph $G'$, Task List $S$, Merged Task $\widetilde{T}$ 
\State Initialize:
Refining granularity hyperparameter $\gamma$.
\State $G' = {\emptyset}$
\While{$G \neq {\emptyset}$}
\State Conduct topological sort on $G$
\State $S = \{ v \in V : \text{indegree}(v) = 0 \}$
\State $\widetilde{T} = \textsc{DynamicMerge}(S, \gamma)$
% \vartriangleleft$ only sub-tasks with release time differences within $\gamma$ will be merged
\State $G' \leftarrow G' \cup \widetilde{T}$
\State $G \leftarrow G \setminus S$
\EndWhile
\State \Return $G'$
\end{algorithmic}
\end{algorithm}

\subsubsection{MIMONet-driven Intermediate Deadline Re-Assignment}
\label{sec:dynamic_runtime_deadline_assignment}

Following intermediate deadline assignment to nodes within a DAG task, the primary objective is to schedule all nodes to meet their deadlines, thereby guaranteeing the fulfillment of the DAG's end-to-end deadline. To achieve this, \tool utilizes the Earliest Deadline First (EDF) scheduling methodology, renowned for its efficacy within both soft and hard real-time system contexts~\cite{48943}. This strategy, based on EDF scheduling, schedules and prioritizes nodes according to their allocated intermediate deadlines. Despite its straightforward nature, this EDF-focused scheduling technique competently addresses major challenges such as dynamic alterations and refinements in workload structure.

Our preliminary studies revealed a distinctive optimization potential in the realm of deadline re-assignment. Specifically, we found that recalculating intermediate deadline assignments at each scheduling point significantly enhances the system's overall average response time performance and minimizes the deadline miss ratio. This is particularly noticeable when a node's execution speed substantially deviates from its estimated execution costs. Furthermore, as illustrated in Figure~\ref{fig:finer_grained_dag}\textbf{(c)}, re-assignment also reduces blocking time\footnote{Blocking time is mainly caused by acquiring a lock on shared memory used by NVIDIA Jetson embedded devices~\cite{bib:nano,bib:tx2,bib:agx,bib:orin}.}, thereby decreasing overall end-to-end latency. Hence, in such circumstances, re-assigning intermediate deadlines can more accurately reflect the real-time workload demand within the system.

Our method dynamically adjusts to the addition or removal of a node from the DAG task by rapidly recalculating the intermediate deadline assignment and scheduling nodes according to their newly assigned deadlines. Despite adding a layer of complexity, this dynamic re-assignment assures optimal resource utilization and increases the probability of adhering to end-to-end deadlines.

\subsection{On-demand Synchronization}
\label{sec:implementation}
Synchronization plays a pivotal role in orchestrating our framework, facilitating effective information sharing throughout the system. However, a naively implemented synchronization strategy can introduce significant overhead, hampering the system's overall performance. Therefore, we leverage an on-demand synchronization strategy to mitigate this issue.

An intuitive, periodic synchronization strategy calls for regular, pre-scheduled synchronization intervals, irrespective of the system's state or needs. This strategy can result in high overhead, as unnecessary synchronization operations may be carried out even when there are no significant changes in the system state. These redundant operations consume valuable computational resources, leading to inefficiencies.

On the other hand, an on-demand synchronization strategy allows for more efficient resource usage. This strategy triggers synchronization only when necessary, i.e., when significant changes in the system state or meeting specific conditions. The synchronization operations are thus tied directly to the system's needs, reducing redundant operations and conserving computational resources. This results in a more responsive system with lower overhead, as illustrated in Figure~\ref{fig:finer_grained_dag}\textbf{(d)}.

\begin{figure}[!t]
\centering
\includegraphics[width=0.5\textwidth]{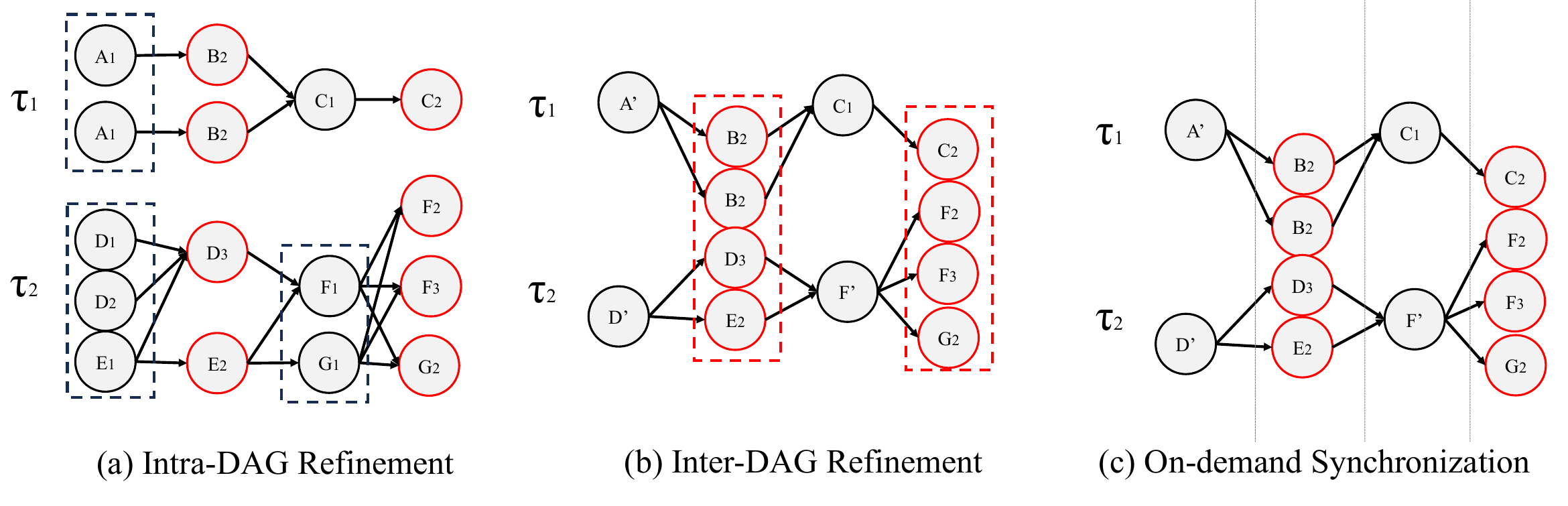}
\caption{A multi-DAG parallel execution example. Intra-DAG refinement aims to merge nodes in the computation graph, thus reducing the computational cost. Inter-DAG refinement increases the parallelism of non-mergeable nodes by batch execution. On-demand synchronization further boosts latency performance by removing unnecessary synchronization overhead.} 
\label{fig:multi_DAG}
\vspace{-2mm}
\end{figure}

\begin{figure}[!t]
\centering
\includegraphics[width=0.4\textwidth]{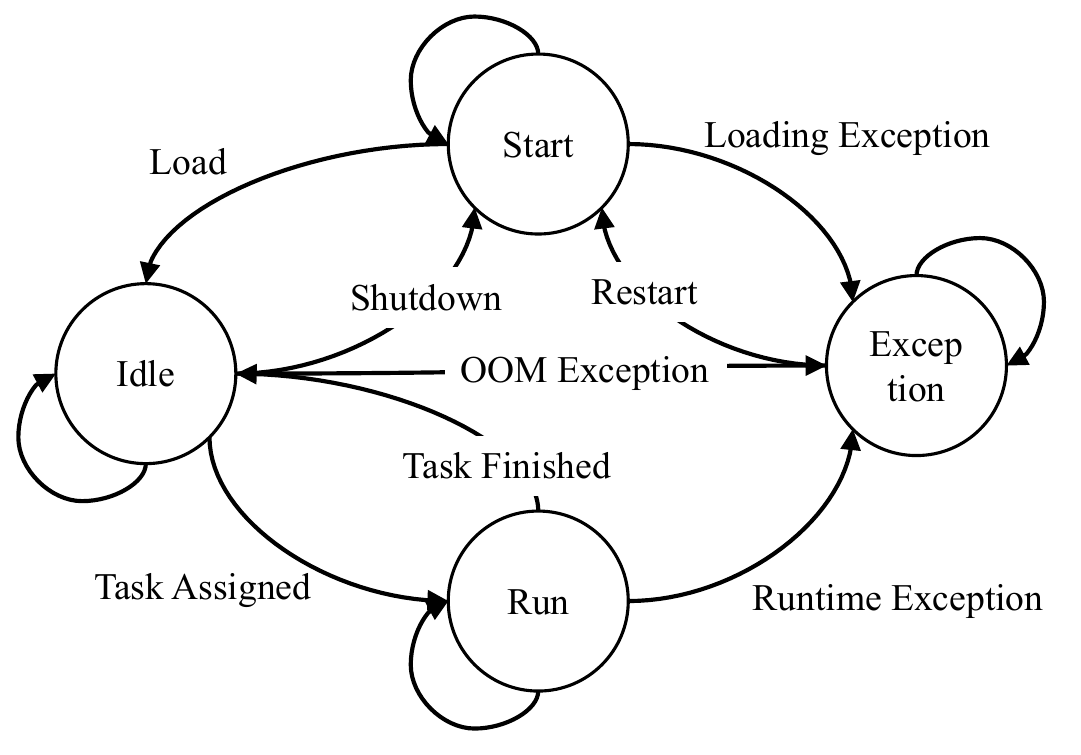}
\caption{Abstract design of finite state machine for handlers. ``OOM'' refers to out-of-memory.}
\label{fig:fsm}
\vspace{-5mm}
\end{figure}

We use a multi-DAG parallel execution scenario to illustrate further combining on-demand synchronization with other components, as in Figure~\ref{fig:multi_DAG}. In this example, within each DAG, we conduct intra-DAG refinement to reduce computation costs by merging some nodes as illustrated in Sec.~\ref{sec:mimonet_case}. Beyond one single DAG, we parallel-execute unmerged independent nodes with the same heights to maximize parallelism. By default, synchronization happens on the edges of nodes. We conduct synchronization only when all the nodes with the same heights finish, reducing the synchronization costs.

\textbf{Global and Local Synchronizers Implementation.} To effectively implement on-demand synchronization in our framework, we introduce a global synchronizer and multiple local synchronizers. The global synchronizer is responsible for system and application-level coordination. Specifically, it periodically invokes the scheduler to dispatch control signals to task handlers, which then request computing resources based on these signals. Upon completion of an inference request, task handlers relay feedback to the scheduler.
At a lower level, each task handler has a dedicated local synchronizer maintaining its internal state. We abstract the handler state into a finite state machine, as depicted in Figure~\ref{fig:fsm}. The local synchronizer periodically reviews the state of this finite state machine. When a state transition occurs, the local synchronizer sends this information to the controller to inform subsequent scheduling decisions.
The global and local synchronizers are implemented based on a spin-lock design, which ensures rapid response times and minimizes overhead, while slightly increasing CPU consumption. This implementation strikes a balance between efficiency and performance, significantly enhancing the system's overall responsiveness and effectiveness. Based on all modules mentioned above, the overall integrated task orchestration algorithm is presented in Algorithm~\ref{alg:integrated}.

\begin{algorithm}[htbp!]
\caption{Integrated Task Orchestration Algorithm}
\label{alg:integrated}
\begin{algorithmic}[1]
\State \textbf{Input:} \
DAG Task $\tau$, \
End-to-end deadline $D$, \
MIMONet Model $M$, \
Timing Requirements $T_{req}$ \
\State \textbf{Output:} \
Refined DAG Task $\tau'$, \
Scheduled Task Execution \
\State $\tau' \leftarrow \textsc{IntermediateDeadlineAssignment}(\tau, D)$
% \Comment{Assign intermediate deadlines}
\State $\tau' \leftarrow \textsc{MIMONetDAGRefinement}(\tau', M)$
\State $\textsc{InitializeSynchronizers}(T_{req})$
\While{not all tasks $\tau_i$ in $\tau'$ are completed}
\State $\textsc{OnDemandSynchronization}()$
% \Comment{Perform on-demand synchronization}
\State $\tau_{next} \leftarrow \textsc{NextTaskToSchedule}(\tau')$
% \Comment{Select the next task for scheduling}
\State $\tau' \leftarrow \textsc{MIMONetDAGReAssignment}(\tau', \tau_{next})$
% \Comment{Re-assign DAG tasks}
\State $\textsc{ExecuteTask}(T_{next})$
\Comment{Execute the scheduled task}
\EndWhile
\end{algorithmic}
\end{algorithm}

% In Algorithm~\ref{alg:integrated}, we present an integrated task orchestration algorithm  that combines intermediate deadline assignment, MIMONet-driven DAG refinement and re-assignment, and on-demand synchronization. This algorithm takes as input a DAG task, an end-to-end deadline, a MIMONet model, and timing requirements. The output is a refined DAG task and the execution schedule for the tasks. First, the algorithm assigns intermediate deadlines to the nodes in the DAG task using the $\textsc{IntermediateDeadlineAssignment}$ function (line 1). Next, it refines the DAG using the MIMONet-driven DAG refinement method with the $\textsc{MIMONetDAGRefinement}$ function (line 2). The synchronizers are initialized with the given timing requirements using the $\textsc{InitializeSynchronizers}$ function (line 3). The algorithm then enters a loop until all tasks in the refined DAG task have been completed (line 4). Within this loop, the algorithm performs on-demand synchronization using the $\textsc{OnDemandSynchronization}$ function (line 5). It then selects the next task to be scheduled using the $\textsc{NextTaskToSchedule}$ function (line 6) and re-assigns the DAG tasks using the $\textsc{MIMONetDAGReAssignment}$ function (line 7). Finally, the selected task is executed with the $\textsc{ExecuteTask}$ function (line 8).

\section{Evaluation}
In this section, we test our full implementation of \tool on top of PyTorch framework~\cite{pytorch} with an extensive set of evaluations. We explore its overall effectiveness under different deadline configurations, from tight to relaxed deadlines, across several hardware platforms with distinct characteristics (Sec.~\ref{sec:overall_effectiveness}). This evaluation aims to demonstrate the general usability and flexibility of \tool in a wide range of scenarios.
Next, we delve into a practical case study, integrating \tool with ROS2 on NVIDIA IoT AI, to understand its effectiveness and adaptability in real-world, complex situations (Sec.~\ref{sec:ros2}). This highlights how seamlessly our solution can be integrated with ROS, an aspect we introduced in the introduction.
Additionally, we conduct a parameter study to demonstrate \tool's flexibility in handling varying deadline settings (Sec.~\ref{sec:adaptability}).
We then evaluate the computational overhead of the \tool framework (Sec.~\ref{sec:overhead}), providing insight into its operational costs.
We conclude the evaluation with a discussion on the outcomes and implications of our analysis (Sec.~\ref{sec:discussion}).

\subsection{Experimental Setups}

\label{sec:setup}

This section details a comprehensive evaluation setup, complete with varying hardware platforms, a diverse set of tasks, and a range of deadline configurations, designed to thoroughly test the \tool framework under a variety of conditions. This breadth of testing helps ensure that \tool can deliver consistent, high-quality performance across many potential use-cases in robotic navigation.

\noindent \textbf{Testbeds.} We use pre-trained models or train DNNs following the original paper's setting on a server with Intel(R) Xeon(R) CPU E5-2650 and a GeForce RTX 2080 Ti GPU. We conduct forward inference experiments on four NVIDIA autonomous embedded platforms as shown in Table~\ref{tab:hardware}. % Specifically, we use NVIDIA Jetson TX2~\cite{bib:tx2}, with 6 big.LITTLE ARM-based cores; NVIDIA Jetson Nano~\cite{bib:nano}, with a Quad-core ARM A57 CPU and 128-core Maxwell-based GPU; and NVIDIA Jetson AGX Xavier~\cite{bib:agx}, an edge computing platform with an 8-core NVIDIA Carmel CPU and a 512-core Volta-based GPU, and NVIDIA Jetson AGX Orin~\cite{bib:orin}, a most recent powerful edge computing platforms equipped with NVIDIA Ampere architecture with 2048 NVIDIA CUDA-core GPU and a 12-core Arm Cortex-A78AE CPU. 
These platforms are widely used in robotics research~\cite{popov2022nvradarnet} applied as the mainboard of various well-known industrial robots, e.g., Duckiebot~\cite{Duckiebot(DB-J)}, SparkFun Jetbot~\cite{SparkFun_JetBot}, Waveshare Jetbot~\cite{Waveshare_JetBot}, etc. 

\begin{table}[!htbp]
  \centering
  \caption{Hardware platforms used in our experiments.}
  \renewcommand\arraystretch{1.3}
  \resizebox{0.49\textwidth}{!}{ 
    \begin{tabular}{|c|c|c|c|c|}
    \hline 
     & {\textbf{Nano}} & {\textbf{TX2}} & {\textbf{Xavier}} & {\textbf{Orin}}\\
    \hline
    {\multirow{3}{*}{CPU}} & 4-core ARM  & 6-core ARM   & 8-core Armv8.2   & 12-core Armv8.2\\
    & Cortex-A57 CPU & Cortex-A57 CPU & Carmel CPU & Cortex-A78AE \\
    & @ 1.43GHz & @ 1.70GHz & @ 2.03GHz & @ 2.20GHz \\
    \hline
    GPU & NV Maxwell GPU & NV Volta GPU & NV Volta GPU & NV Ampere GPU \\
    \hline
    Memory & 4GB LPDDR4 & 8GB LPDDR4 & 16GB LPDDR4x & 32GB LPDDR5 \\
    \hline
    Storage & 16GB eMMC 5.1 & 32GB eMMC 5.1 & 32GB eMMC 5.1 & 64GB eMMC 5.1 \\
    \hline
    \end{tabular}
}
  \label{tab:hardware}
  \vspace{-3mm}
\end{table}

\noindent \textbf{Metrics.}
We evaluate our system's performance primarily through three critical metrics: latency, deadline miss rate, and Quality of Experience (QoE) score. Latency is the time taken for a request to be processed and returned, with lower times indicating faster responses. The deadline miss rate is the proportion of tasks that fail to meet their specified deadlines, which we aim to minimize. Lastly, we define the QoE score (detailed in Eq.\ref{eq:qoe}) motivated by~\cite{kwon2022xrbench}, providing a comprehensive measure of user satisfaction in soft real-time systems by considering aspects such as responsiveness, reliability, and overall service quality.

\begin{equation}
QoEScore(r) = \frac{1}{1 + e^{\lambda}[max(0, C_r-S_r)]}
\label{eq:qoe}
\end{equation}

Here, $C_r$ represents the execution time of the inference task $r$, and $S_r$ is the slack of task $r$. $\lambda$ is a hyper-parameter controlling latency tolerance. This metric refers to the exponential distribution, and for this study, we set $\lambda$ to 1. By closely monitoring and optimizing these metrics, we ensure a seamless and efficient user experience.

\noindent \textbf{Baselines.}
We implement the \tool scheduling framework (refer Sec.~\ref{sec:overview}), which leverages a single MIMONet model to adhere to memory constraints in robotic embedded deployment. We compare \tool against these baselines\footnote{Recall that most of the existing DAG scheduling methods could not handle dynamically changing workloads and thus are not applicable to our problem scope. To our knowledge, one feasible solution to handle the dynamically changing workload scenario is the classical intermediate deadline assignment method designed for scheduling DAGs~\cite{panahi2009framework, liu2000real}, which we evaluated as one of the baselines.}: (1) \textbf{EDF}~\cite{liu2000real}: a strict version of the EDF scheduler; (2) \textbf{\tool-FG}: fine-grained MIMONet partitioning with proportional deadline assignment~\cite{panahi2009framework} following EDF scheduling; (3) \textbf{\tool-IDA}: an optimized intermediate deadline assignment (IDA) based on \tool-FG serving as a robust baseline; and (4) \textbf{\tool}: the comprehensive \tool implementation with fine-grained DAG partitioning, optimized IDA, and an on-demand synchronization mechanism.

\noindent \textbf{Minibenchmark.}
To examine the temporal correctness and throughput performance of the scheduling approach in a dynamic robotic environment, we establish a mini-benchmark employing deep learning tasks as in Table~\ref{tab:models}. As detailed in Table~\ref{tab:minibench}, we configure a sequential three-stage task setting for both environments, each predicated on completing the previous stage. The end-to-end task involves sequential execution of the obstacle-free cruising task set and obstacle cruising task set. Each task set is executed 10 times in the benchmark. The end-to-end deadline for each deep learning task instance, based on the platform-wise worst-case execution time (WCET), is set under two configurations: tight and loose. For our mini benchmark, the tight and loose deadlines were established as 9815ms and 11325ms for Nano; 8400ms and 10080ms for TX2; 6500ms and 7315ms for Xavier; and 2400ms and 3600ms for Orin. We set hyperparameter $\gamma$ as 100ms. 
Note that for almost all these benchmark workloads, we observe that  GPU utilization reaches nearly 100\% consistently. This implies that modern DNN inference is extremely GPU-intensive and can naturally exploit the maximum parallelism provided by GPU hardware employed on most embedded devices.

\begin{table}[!htbp]
\centering
\caption{Minibenchmark design for evaluating multi-DNN based robots in dynamic environments. ``L'', ``S'', ``C'', and ``O'' refer to lane detection, segmentation, cruise control, and object detection, respectively.}
\resizebox{0.45\textwidth}{!}{
\begin{tabular}{|c|ccc|}
\hline
& \textbf{Stage1} & \textbf{Stage2} & \textbf{Stage3} \\
\hline
Obstacle-free cruising & L & S & C \\
Obstacle cruising & L & S+O & C \\
\hline
\end{tabular}
}
\label{tab:minibench}
\end{table}

\begin{figure*}[!t]
    \centering
    \includegraphics[width=\textwidth]{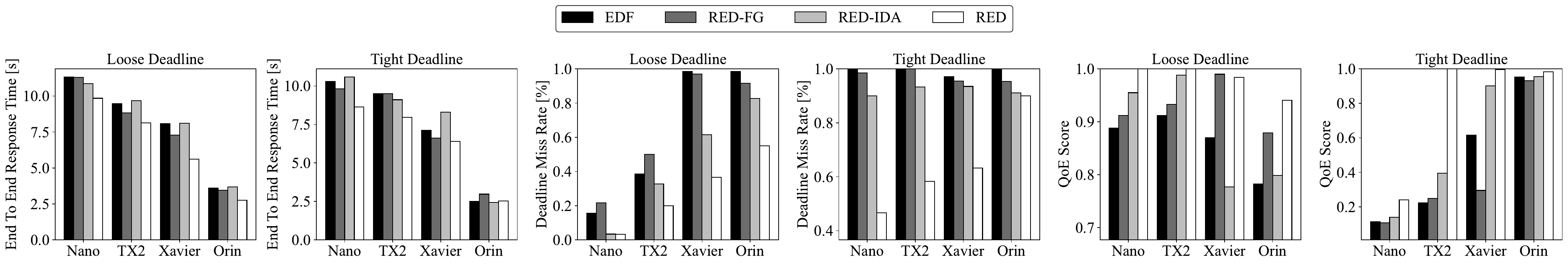}
    \caption{Overall effectiveness of \tool evaluated on four different resource-constrained intelligent robotic systems. \tool optimizes throughput, real-time performance, and QoE, demonstrating substantial improvements over baseline methods. }
    \label{fig:overall_effectiveness}
    \vspace{-5mm}
\end{figure*}

\begin{figure}[!t]
    \centering
    \includegraphics[width=0.48\textwidth]{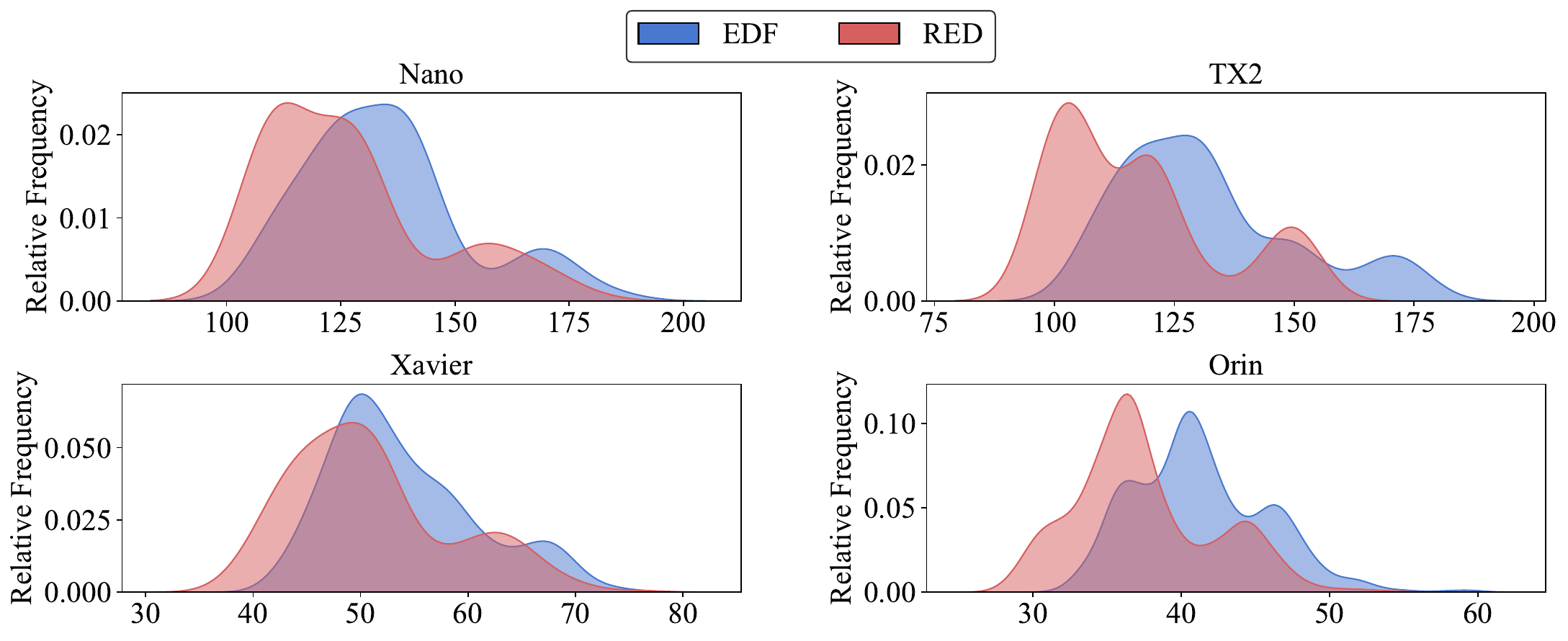}
    \caption{Response time histogram of \tool across different embedded platforms compared to EDF.}
    \label{fig:hist}
    \vspace{-5mm}
\end{figure}

\subsection{Overall Effectiveness}
\label{sec:overall_effectiveness}

% \zx{add one latency distribution figure like this for latency histograms if we have data?? @Tao}

In this section, we conduct a comprehensive evaluation study of \tool, taking into account all its components. Our design aims to optimize system throughput, real-time performance, and quality of experience (QoE).

\noindent \textbf{Throughput.}
As depicted in Figure~\ref{fig:overall_effectiveness}, \tool consistently outperforms the baseline approaches under all variant deadline settings. The average latency is reduced by 24.7\%, 17.3\%, and 14.2\% compared to EDF, \tool-IDA, and \tool-FG, respectively. These results demonstrate the overall throughput effectiveness of \tool. Furthermore, \tool-FG and \tool-IDA both outperform EDF in seven out of eight settings. This further illustrates that each subcomponent of our design could almost always result in throughput performance improvements. Delving into the details, we observe that \tool-FG and \tool-IDA significantly outperform EDF in all loose deadline settings. However, under tight deadline settings, they outperform EDF in three out of four cases correspondingly. This observation suggests that while the subcomponents of \tool can improve throughput in most cases, their effectiveness might be more pronounced in scenarios with less stringent deadlines. The reduced effectiveness under tight deadline settings could be due to the increased scheduling complexity or the limited optimization opportunities in such constrained environments. % Further investigation and refinement of the subcomponents may help enhance their performance under tight deadline conditions. 
To further demonstrate the effectiveness of our proposed \tool, we show the histogram of task-level response time in the case of tight and loose deadline setting, respectively. As depicted in Figure~\ref{fig:hist}, we plot the response time histogram of the proposed \tool compared with EDF.  \tool's holistic efficient design offers significantly faster responses qualtitavely. 

\noindent \textbf{Real-time Performance.} Figure~\ref{fig:overall_effectiveness} illustrates the deadline miss rate for different approaches tested on four embedded devices. The results reveal that our proposed \tool achieves superior real-time performance, surpassing the baselines considerably. Specifically, \tool significantly outperforms the baseline approaches under nearly all settings, with an average deadline miss rate lower than that of EDF, \tool-IDA, and \tool-FG by 40.5\%, 37.8\%, and 37.5\%, respectively. Interestingly, we observe that merely applying finer-grained DAG partitioning (\tool-FG) and further implementing optimized intermediate deadline assignments (\tool-IDA) results in a marginal improvement over the EDF baseline, with increases of 4.8\% and 4.4\%, respectively. This marginal improvement can be attributed to various factors, one of the most significant being the high synchronization overhead. \tool's on-demand synchronization mechanism effectively reduces this overhead, leading to markedly better real-time performance.

\noindent \textbf{Quality of Service.} In the embedded  system scenario, imposing a stringent deadline on the heavy MIMONet workload often results in missed deadlines, as shown in Figure~\ref{fig:overall_effectiveness}. This section investigates the potential of \tool in enhancing the quality of experience (QoE) across various settings, following the methodology outlined in~\cite{kwon2022xrbench}. Figure~\ref{fig:overall_effectiveness} presents the QoE for different approaches evaluated on four embedded devices. The findings demonstrate that \tool consistently achieves superior QoE, significantly outperforming the baseline methods. In particular, \tool surpasses the QoE scores of EDF, \tool-IDA, and \tool-FG on average by 34.8\%, 34.2\%, and 15.9\%, respectively. In the loose deadline setting, \tool exceeds others by a relatively smaller margin on average (13.1\%, 9.8\%, and 1.4\%). Conversely, in the tight deadline setting, the baseline candidates exhibit lower QoE, with \tool significantly outperforming them on average by 76.4\%, 84.7\%, and 40.4\%. Notably, in specific scenarios such as TX2-tight, EDF's QoE score is a mere 0.222, while \tool achieves a QoE of 1.0, which is 3.50 times higher than the EDF baseline.
In summary, \tool consistently delivers superior QoE across various settings in MIMONet workload systems, outperforming baseline methods by substantial margins, particularly under tight deadline constraints. \\

\begin{minipage}{0.45\textwidth}
\begin{shaded}
    \noindent{\textbf{On-device overall effectiveness}}: \tool effectively addresses all challenges outlined in Sec.\ref{sec:motivation} for resource-constrained intelligent robotic systems. It optimizes throughput, real-time performance, and QoE in MIMONet workload systems, demonstrating substantial improvements over baseline methods, especially under tight deadline constraints.
\end{shaded}
\end{minipage}

\subsection{A Practical Case Study on ROS2}
\label{sec:ros2}

We implemented a practical case study on the Robot Operating System 2 (ROS2) utilizing the open-source NVIDIA IoT AI framework~\cite{nvidiaiot}. Specifically, we focused on a realistic robotic application involving a camera input with multiple inference outputs, including lane detection, segmentation, cruise control, and object detection. And we comprehensively evaluate the effect of \tool working under different GPU interference intensities.

\begin{figure}[!t]
\centering
\includegraphics[width=0.5\textwidth]{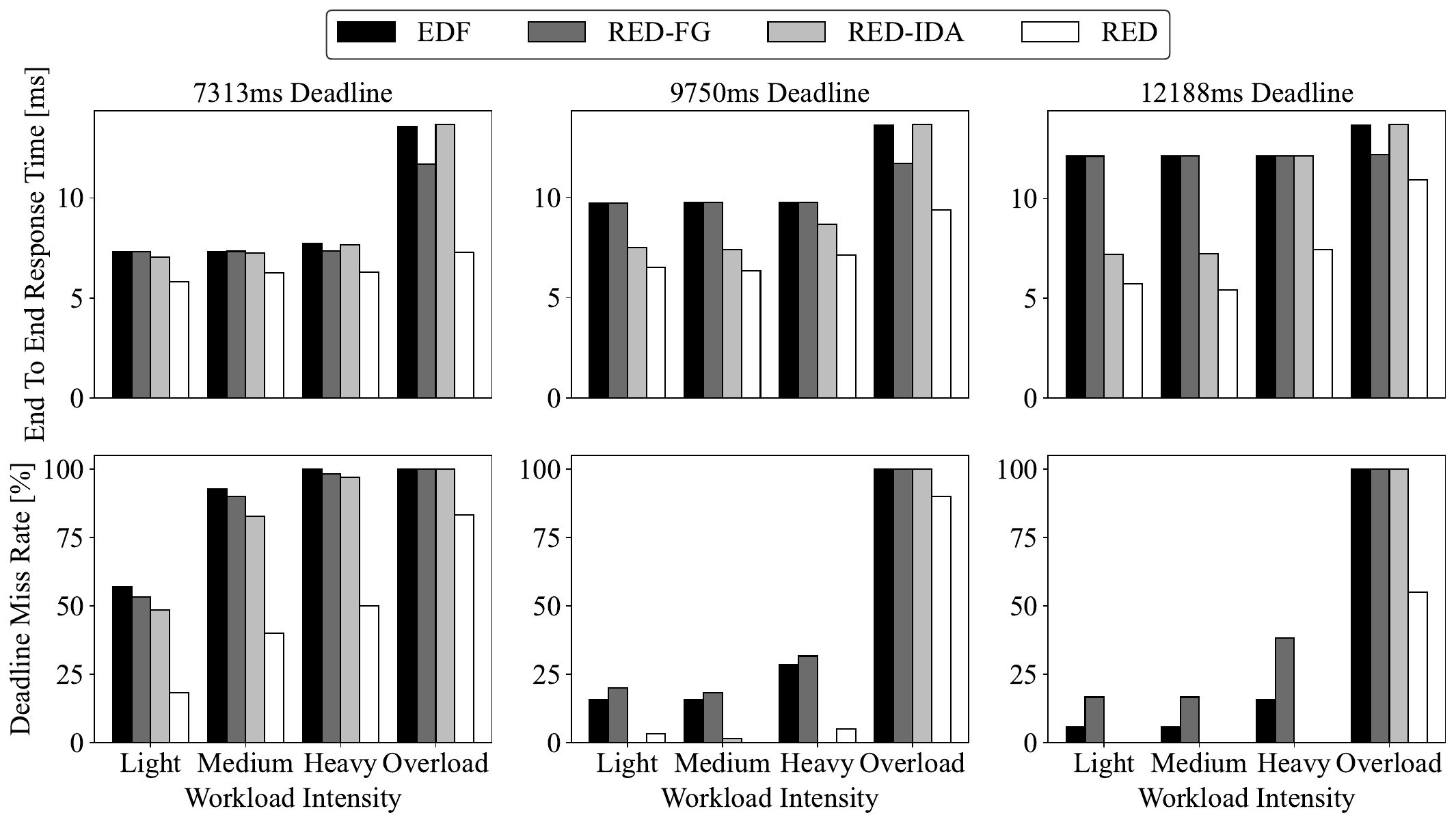}
\caption{A practical case study on ROS2 involving a camera input with multiple inference outputs on Xavier.}
\label{fig:ros2_example}
\vspace{-3mm}
\end{figure}

To support MIMONet's inference, we modified the open-source library provided by NVIDIA. As shown in Figure~\ref{fig:ros2_example}, unlike traditional single-task DNNs requiring deployment across multiple ROS nodes, we deployed the MIMONet model on a single ROS node. This modification significantly reduced the communication overhead intrinsic to ROS while enhancing the controllability of each module.
Additionally, we simulate a camera mounted on a car, abstract as a node in ROS2. The MIMONet model processed this data to produce outputs for several deep learning models, as mentioned before. To conduct a fair comparison, we fully implement EDF, RED, and its ablations built on ROS2, strictly following our design.

As depicted in Fig.~\ref{fig:ros2_example}. Quantitative results from this case study demonstrated the superior performance of the MIMONet model deployed on a single ROS node compared to traditional single-task DNNs deployed on multiple nodes. Compared to the EDF scheduler built upon ROS2, \tool provides consistently lower deadline miss rates on average by 67.3\% and 32.7\% lower response time.

% vides 32.7\% less response time, lower deadline miss rates by 67.3\%. 

These significant improvements underline the practical usability of MIMONet models in integrating with real-world ROS2 systems. We emphasize that in most interference scenarios, \tool integrated with ROS2 consistently has better timing performance than all the ablations and baseline EDF 
by a large margin, further evidencing the practical usability of \tool.

% almost no missed deadline. even in the extreme case of malicious workload, where EDF misses all deadlines, \tool can keep the deadline miss rate low. \\

\vspace{2mm}
\begin{minipage}{0.45\textwidth}
\begin{shaded}
\noindent{\textbf{Practical Usability}}: This case study demonstrates the practical deployment of \Approach{} on ROS2-based robotic applications. \Approach{} significantly reduces latency and improves deadline compliance, highlighting the usability of \Approach{} in complex robotic navigation scenarios.
\end{shaded}
\end{minipage}

% \begin{center}
% \begin{tcolorbox}[colback=gray!10,
%                   colframe=black,
%                   width=0.49\textwidth,
%                   arc=1mm, auto outer arc,
%                   boxrule=0.9pt,
%                  ]
% \textit{\textbf{Practical Usability}}: This case study demonstrates the practical deployment of \Approach{} on ROS2-based robotic applications. \Approach{} significantly reduces latency and improves deadline compliance, highlighting the usability of \Approach{} in complex robotic navigation scenarios.
% \end{tcolorbox}
% \end{center}

\subsection{Adaptability under Different Scenarios}
\label{sec:adaptability}

\noindent \textbf{Adaptability to variant deadlines.} To investigate the adaptability of the \tool system to various end-to-end deadlines, we conducted a parameter study on the NVIDIA Jetson AGX Xavier platform. This study evaluates the system's performance under different deadlines, ranging from tight to loose. In real-world scenarios, end-to-end deadlines can change due to environmental dynamics, and a robust system should adapt to these changes to meet the required deadlines.
As depicted in Table.~\ref{fig:variant_deadline}, \tool outperforms the baseline approaches under all deadline settings, with an average latency less than EDF, \tool-IDA, and \tool-FG, on average by 23.2\%, 20.1\%, and 17.1\%, respectively. This suggests that \tool can adapt to varying end-to-end deadlines by applying a MIMONet-aware intermediate deadline assignment policy integrated with dynamic DAG reconstruction, even in the face of environmental dynamics.
Moreover, by incorporating components of \tool, the end-to-end response time gradually decreases, significantly outperforming the EDF baseline. This observation further supports the idea that each component of our proposed \tool contributes to increasing throughput and fully utilizing system resources to achieve a fast response time.

\noindent \textbf{Adaptability to variant QoE requirements.} To assess the versatility of our tool in the context of diverse QoE requirements, we maintained the workload intensity at a constant level while adjusting the hyperparameter $\lambda$ of the QoE metrics, as defined in Eq.~\ref{eq:qoe}. This experiment was performed across four unique embedded devices using a stringent tight deadline benchmark as defined in Sec.~\ref{sec:setup}.
As depicted in Table~\ref{fig:variant_lambda}, our \tool consistently outperforms the alternatives across an extensive range of hyperparameter $\lambda$ values, varying from 0.001 to 10. This result holds true under constant workload conditions, demonstrating the adaptability of our tool to different system settings.

\begin{table}[!htbp]
\centering
\caption{End-to-end latency of different approaches under
variant deadline settings on Xavier. A larger deadline-setting ID indicates a looser deadline. The best values are in bold.}
\resizebox{0.47\textwidth}{!}{
\begin{tabular}{c|cccccc}
\hline
Deadline Setting ID & 1 & 2 & 3 & 4 & 5 & 6 \\
\hline
EDF & 5869 & 6807 & 6865 & 7760 & 8737 & 9694 \\
RED-FG & 5883 & 6006 & 6803 & 7856 & 8723 & 9688 \\
RED-IDA & 4950 & 4929 & 6691 & 7465 & 8313 & 9045 \\
RED & \textbf{4455} & \textbf{4435} & \textbf{4456} & \textbf{4500} & \textbf{5930} & \textbf{6779} \\
\hline
\end{tabular}}
\label{fig:variant_deadline}
% \vspace{-2mm}
\end{table}

\begin{table}[!htbp]
\centering
\caption{QoE scores of different approaches under variant hyperparameter settings across four embedded platforms. Larger $\lambda$ indicates higher QoE requirements. The best values are in bold.}
\resizebox{0.47\textwidth}{!}{
\begin{tabular}{c|c|cccccc}
\hline
& $\lambda$ Value & 0.001 & 0.01 & 0.1 & 1.0 & 10.0 \\
\hline
\multirow{4}{*}{Nano} & EDF & 0.256 & 0.254 & 0.238 & 0.113 & 0.001 \\
& RED-FG & 0.248 & 0.246 & 0.230 & 0.108 & 0.001 \\
& RED-IDA & 0.306 & 0.304 & 0.285 & 0.140 & 0.002 \\
& RED & \textbf{0.461} & \textbf{0.460} & \textbf{0.437} & \textbf{0.240} & \textbf{0.004}\\
\hline
\multirow{4}{*}{TX2} & EDF & 0.420 & 0.418 & 0.399 & 0.222 & 0.003 \\
& RED-FG & 0.473 & 0.470 & 0.448 & 0.248 & 0.001 \\
& RED-IDA & 0.638 & 0.636 & 0.615 & 0.395 & 0.001 \\
& RED & \textbf{1.000} & \textbf{1.000} & \textbf{1.000} & \textbf{1.000} & \textbf{1.000}\\
\hline
\multirow{4}{*}{Xavier} & EDF & 0.748 & 0.746 & 0.735 & 0.616 & 0.003 \\
& RED-FG & 0.525 & 0.523 & 0.501 & 0.295 & 0.005 \\
& RED-IDA & 0.946 & 0.945 & 0.942 & 0.902 & 0.500 \\
& RED & \textbf{0.999} & \textbf{0.999} & \textbf{0.998} & \textbf{0.996} & \textbf{0.900}\\
\hline
\multirow{4}{*}{Orin} & EDF & 0.973 & 0.972 & 0.971 & 0.954 & 0.878 \\
& RED-FG & 0.953 & 0.952 & 0.951 & 0.931 & 0.893 \\
& RED-IDA & 0.973 & 0.973 & {0.971} & {0.956} & {0.900}\\
& RED & \textbf{0.990} & \textbf{0.990} & \textbf{0.989} & \textbf{0.984} & \textbf{0.960} \\
\hline
\end{tabular}}
\label{fig:variant_lambda}
\vspace{-2mm}
\end{table}

\begin{minipage}{0.45\textwidth}
\begin{shaded}
\noindent{\textbf{Robust Adaptability}}: \tool demonstrates exceptional adaptability under varied end-to-end deadlines and QoE requirements. It consistently outperforms baselines across multiple platforms and scenarios, showcasing its robustness and efficacy in dynamic environments.
\end{shaded}
\end{minipage}

\subsection{Overhead Analysis}
\label{sec:overhead}

\noindent \textbf{Memory Overhead.}
Our method demonstrates exceptional memory efficiency, with minimal overhead, as detailed in Table~\ref{tab:memory_overhead}. As expected, our highly efficient implementation ensures low memory overhead values for both the scheduler and synchronization components, with the scheduler being particularly modest.
However, it is noteworthy that the on-demand synchronization components slightly compromise memory efficiency to enhance latency through reduced context switching between processes. As discussed in Sec.~\ref{sec:implementation}, on-demand synchronization is implemented to decrease code design space coupling, thereby increasing reusability. This approach leads to increased memory overhead due to complex inter-process communication. Future work may consider a potentially more memory-efficient alternative, such as implementing our solution in a multithreading manner, though this may trade off implementation complexity and reusability.
Notably, the overhead ratios consistently remain low across four different hardware platforms and configurations, evidencing the memory efficiency of our approach.

\begin{table}[!htbp]
\centering
\caption{Detailed Memory overhead and percentage of applying \Approach{} for each minibenchmark in this paper. }
\resizebox{0.48\textwidth}{!}{
\begin{tabular}{c|cc|cccc}
\hline
& \multicolumn{2}{c|}{(a) Overhead Raw Value} & \multicolumn{4}{c}{(b) Overhead Ratio}\\
\hline
& Scheduler & Synchronization & Nano & TX2 & Xavier & Orin \\
\hline
Tight & 5 KB & 40.1 MB & 1.00\% & 0.50\% & 0.25\% & 0.13\% \\
Loose & 3 KB & 36.9 MB & 0.92\% & 0.46\% & 0.23\% & 0.12\% \\
\hline
\end{tabular}}
\label{tab:memory_overhead}
% \vspace{-2mm}
\end{table}

\begin{table}[!htbp]
\centering
\caption{Average runtime overhead of \Approach{} are shown as the first row. Profiling overhead is shown as the second row.}
\resizebox{0.4\textwidth}{!}{
\begin{tabular}{c|cccc}
\hline
& Nano & TX2 & Xavier & Orin \\
\hline
Runtime Overhead (ms) & 27.3 & 26.9 & 17.4 & 1.2 \\
Profiling Overhead (s) & 51.1 & 34.3 & 19.1 & 16.1 \\
\hline
\end{tabular}%
}
\label{tab:overhead_exec}%
\vspace{-2mm}
\end{table}

\noindent \textbf{Execution Overhead.}
In addition to memory overhead, the execution overhead of our proposed method is also critical. Table~\ref{tab:overhead_exec} shows the runtime execution overhead of \tool and the profiling overhead for each testing platform. The first line of table showcases the remarkably low runtime execution overhead due to the efficient implementation of the \Approach{} framework. The scheduling overhead for NVIDIA Jetson Orin is significantly lower, potentially due to its utilization of a newer Linux kernel version. The main source of execution time overhead arises from scheduling decision-making. One potential strategy to reduce such overheads in future work involves employing finer-grained locks, especially when adapting to more stringent resource-constrained scenarios. Compared to latency data, the runtime overhead remains reasonably low, further underscoring the effectiveness of our approach.
The second line of the table examines the profiling overhead of \tool. The offline profile overhead is deemed acceptable, even on the Jetson Nano, which displays the largest absolute overhead value of merely 51.1 seconds (approximately 1 minute). Importantly, for a fixed system configuration, the offline profile requires only a one-time effort, thus easing its integration into continuous integration and deployment of \tool.

\begin{minipage}{0.45\textwidth}
\begin{shaded}
\noindent\textbf{Low overhead}: Our method demonstrates not only impressive memory efficiency but also low execution overhead across various platforms. The efficient implementation of the \Approach{} framework, as well as the modest overhead incurred by the scheduler and synchronization components, contribute significantly to these outcomes. 
\end{shaded}
\end{minipage}

% \clearpage

\subsection{Discussion}
\label{sec:discussion}
\noindent \textbf{Solution Scalability.} The scalability of our proposed solution is primarily hindered by the computational constraints of the devices in use. In instances where training the MIMONet from scratch is impractical by limited computing resources, leveraging pre-trained models can significantly alleviate this issue. These models, often trained on extensive datasets, can serve as efficient encoders for the MIMONet model, thereby greatly reducing the required training time. For instance, state-of-the-art pre-trained visual transformer (ViT) models on the ImageNet dataset~\cite{deng2009imagenet} which consist of over 14M images can be utilized as shared encoders for navigation robot tasks. Additionally, the low-coupling design of the \tool's controller and handler modules offers flexibility, allowing the framework to be conveniently modified for decentralized deployment through the integration of a network communication module.

\noindent \textbf{Future Work.} Looking ahead, there are several promising directions for enhancing the functionality and adaptability of our framework. Beyond the weight-sharing approach employed in this work, integrating existing strategies for deploying deep learning models on embedded devices, such as model compression~\cite{VIB,li2023mimonet}, pruning, and quantization, could further optimize our system. These enhancements could be designed to be both configurable and reusable. On the application front, integrating state-of-the-art structures, such as Transformers-based large language models (LLMs)~\cite{zhang2022investigating,chen2021revisiting,chen2022generate,DBLP:conf/acl/LiLGL23,DBLP:conf/acl/ChenCL0LT023}, into \tool could potentially enhance accuracy. In terms of device compatibility, the landscape of AI-oriented embedded devices is rapidly evolving, with many now equipped with function-specific computational devices like Field-Programmable Gate Arrays (FPGAs) and Neural Processing Units (NPUs). This trend towards heterogeneous Systems-on-a-Chip (SoCs) suggests another future direction: optimizing \tool's strategy to account for the unique characteristics of these heterogeneous computing devices with integrating processor scheduling~\cite{wang2022towards,li2021efficient}. This enhancement would expand the range of embedded devices that can deploy \tool on-device.

\section{Related Work}
\noindent\textbf{Real-time DAG scheduler.} DAG-based workloads have received much attention in both server systems~\cite{ahmed2022exact,sakellariou2004hybrid,wu2015hierarchical,canon2008comparative,panahi2009framework} and embedded systems~\cite{xie2015heterogeneity,bi2022response}. Theoretical multi-processor DAG scheduling has been explored dating back to Graham's bound in 1969~\cite{graham1969bounds}.
Recently, more advancements have been achieved in real-time DAG scheduling and analysis of parallel workloads~\cite{ueter2018reservation,baruah2015federated,jiang2020real,jiang2017semi,li2014analysis,li2013outstanding,melani2015response,bi2022response,he2022bounding}. However, to our knowledge, most existing DAG schedulers cannot handle MIMONet and dynamically changing workloads, thus being  in-applicable to our problem. A classical real-time DAG scheduling approach, the intermediate deadline assignment~\cite{panahi2009framework, liu2000real}, can be leveraged to handle dynamically changing workloads, which we evaluated as one of the baselines.

\noindent\textbf{Real-time ROS scheduler.} 
Several real-time schedulers have been proposed to handle various tasks in the Robot Operating System (ROS) environment in recent years, encompassing dynamic priority scheduling, latency optimization, feedback-based scheduling, multicore awareness, and resource-aware scheduling.~\cite{DBLP:conf/rtss/JiangJGLTW22,DBLP:conf/rtss/LiGJGDL22,DBLP:conf/rtss/TeperGUBC22,DBLP:conf/rtss/BlassCBB21,DBLP:conf/rtss/TangFG0LD020,DBLP:conf/rtas/ChoiXK21,DBLP:conf/rtas/BlassHLZB21}
Although prior research has significantly advanced ROS-based real-time scheduling, these works mainly target traditional tasks, not accounting for the unique demands of Multiple-Input Multiple-Output Deep Neural Networks (MIMONet). Our work distinctively addresses this gap, proposing an innovative scheduler specifically designed for MIMONet workloads, thereby catering to the unique requirements of real-time DNN execution in a ROS environment.

\noindent\textbf{Real-time DNN Inference.}
Recent advancements in real-time DNN research have significantly improved the performance of low-latency solutions. Various strategies and frameworks have been developed to optimize the trade-off between performance and accuracy. For instance, dynamic approximation strategies, supervised streaming and scheduling frameworks, and energy-efficient execution approaches have been proposed to boost real-time performance~\cite{jeong2022band,bateni2018apnet, zhou2018s, predjoule}. Also, some research efforts have focused on creating operating systems specifically designed for DNN execution, exploring theoretical schedulability, and developing methods for the multi-DNN inference that maximize the use of available CPU and GPU resources~\cite{neuos, kang2021lalarand, xiang2019pipelined}. However, despite these significant strides, to our knowledge, a research gap persists in the domain of real-time systems for Multiple-Input Multiple-Output (MIMO) DNNs. This paper addresses this gap by formulating and resolving corresponding problems in this sphere.

\noindent\textbf{Weight-Shared DNNs.}
Weight-sharing techniques enhance the feasibility of deploying Deep Neural Networks (DNNs) on low-memory devices, applicable in both single-task~\cite{bib:ICLR19:Yu,bib:ICML17:Bolukbas,bib:MobiCom18:Fang} and multi-task setups~\cite{bib:IJCAI18:Chou,bib:NIPS18:He, bib:TMC21:He, bib:MobiSys20:Lee, bib:APSIPA20:Wu}. In the former, different DNN variants provide varied workload-accuracy trade-offs for the same task, whereas in the latter, parameter-sharing occurs among related tasks, achieved through methods like cross-model quantization or fine-tuning.

\noindent\textbf{Intelligent Robotic Systems}
Recent intelligent robotic systems are expected to be capable of multi-tasking, corresponding to deep learning models. Several researchers focus on embedded training~\cite{yoo2022learning} or inference~\cite{lv2022sagci,popov2022nvradarnet} of deep learning models in robotic scenarios. In both scenarios, the intelligent robotic systems are supposed to be efficient, light-weighted, and mostly required to ensure timing correctness under soft or hard real-time constraints for safety-critical or strongly interactive scenarios such as multi-robot communication~\cite{nikkhoo2023pimbot,he2023robust}, UAV trajectory planning~\cite{xu2022dpmpc}, geo-localization~\cite{zhang2022learning}, robotic tracking~\cite{botros2022fully}, human-computer interaction~\cite{maccio2022mixed,lee2022towards}, autonomous driving~\cite{he2022robust,he2023robustev}, etc.

\section{Conclusion}
In this paper, we introduce \tool, a comprehensive framework tailored for multi-task DNN inference on resource-restricted robotic systems, designed to adaptively navigate \textbf{R}obotic \textbf{E}nvironmental \textbf{D}ynamics under real-time constraints. Central to \tool is a deadline-driven scheduler incorporating an intermediate deadline assignment policy, adept at managing evolving workloads and asynchronous inference amidst unpredictable environments. Furthermore, \tool effectively supports the deployment of MIMONet (multi-input multi-output neural networks), addressing memory limitations and exploiting the unique weight-sharing architecture of MIMONet. Consequently, \tool devises an innovative workload refinement and reconstruction process, enabling compatibility with MIMONet and optimizing efficiency. The comprehensive evaluation of \tool underscores its effectiveness concerning throughput, timing correctness, practical usability, adaptability, and low overhead. 
However, it's worth noting that the current design might be best suited for a specific DNN structure (i.e., MIMONet). Our implemented DAG scheduling algorithm, in its present form, is relatively simple and practical. We plan to leverage interesting and relevant ideas from many real-time DAG scheduling approaches proposed recently and further enhance the performance of \Approach{} in the near future.

\section*{Acknowledgments}
This research was supported by the National Science Foundation under Grants CNS Career 2230968, CPS 2230969, CNS 2300525, CNS 2343653, CNS 2312397.

\bibliographystyle{IEEEtran}
\bibliography{rtss23-red}

\end{document}